\documentclass[conference]{IEEEtran}
\IEEEoverridecommandlockouts
\usepackage{cite}
\usepackage{enumitem}
\usepackage{amsmath,amssymb,amsfonts}
\usepackage{algorithmic}
\usepackage{graphicx}
\usepackage{textcomp}
\usepackage{xcolor}
\usepackage{tikz}
\usepackage[algo2e,ruled,linesnumbered,boxed,commentsnumbered]{algorithm2e}
\usepackage{latexsym}
\usepackage{dcolumn}
\usepackage{booktabs}
\usepackage{amsmath}
\usepackage{multirow}
\usepackage{graphicx}
\usepackage{amssymb}
\usepackage{pifont}
\newcommand{\cmark}{\ding{51}}%
\newcommand{\xmark}{\ding{55}}%
\usepackage{amsmath,array,graphicx}
\usepackage{kantlipsum}
\def\BibTeX{{\rm B\kern-.05em{\sc i\kern-.025em b}\kern-.08em
    T\kern-.1667em\lower.7ex\hbox{E}\kern-.125emX}}
    
\usepackage{url}

\usepackage{breakurl}
\usepackage[breaklinks]{hyperref}
\usepackage{rotating}
    
\begin{document}

\title{Communication Efficiency in Federated Learning: Achievements and Challenges}

\author{
    \IEEEauthorblockN{Osama Shahid\IEEEauthorrefmark{1}, Seyedamin Pouriyeh\IEEEauthorrefmark{1}, Reza M. Parizi\IEEEauthorrefmark{2}, Quan Z. Sheng\IEEEauthorrefmark{3}, Gautam Srivastava\IEEEauthorrefmark{4},  Liang Zhao\IEEEauthorrefmark{1}}
    \IEEEauthorblockA{\IEEEauthorrefmark{1} Department of Information Technology, Kennesaw State University, Marietta, GA, USA
    \\oshahid@students.kennesaw.edu, \{spouriye, lzhao10\}@kennesaw.edu}
    \IEEEauthorblockA{\IEEEauthorrefmark{2} Department of Software Engineering and Game Development, Kennesaw State University, Marietta, GA, USA   \\rparizi1@kennesaw.edu}

   \IEEEauthorblockA{\IEEEauthorrefmark{3}Department of Computing, Macquarie University, Sydney, Australia
    \\ michael.sheng@mq.edu.au }
       \IEEEauthorblockA{\IEEEauthorrefmark{4}Department of Math and Computer Science, Brandon University, Canada
    \\  }
    
         \IEEEauthorblockA{\IEEEauthorrefmark{4}Research Centre for Interneural Computing, China Medical University, Taichung Taiwan
    \\ srivastavag@brandonu.ca}

 \thanks{ \scriptsize \noindent Corresponding author: S. Pouriyeh (email: spouriye@kennesaw.edu).}   
}

\maketitle

\begin{abstract}
Federated Learning (FL) is known to perform Machine Learning tasks in a distributed manner. Over the years, this has become an emerging technology especially with various data protection and privacy policies being imposed FL allows performing machine learning tasks whilst adhering to these challenges. As with the emerging of any new technology, there are going to be challenges and benefits. A challenge that exists in FL is the communication costs, as FL takes place in a distributed environment where devices connected over the network have to constantly share their updates this can create a communication bottleneck. In this paper, we present a survey of the research that is performed to overcome the communication constraints in an FL setting.
\end{abstract}


\section{Introduction}
A Machine Learning (ML) model for a specific task is created utilizing the unprecedented amount of data that is generated from devices. This data can be used to achieve process optimization \cite{weichert2019review}, gain insight discovery \cite{pazzani1997comprehensible} and aid in decision making \cite{meyer2014machine}. Some examples of devices that generated data can be smartphones and wearable devices, smart homes, etc. Traditionally, to implement ML predictive models the data would need to be transferred to a central server where it could be stored and used for training and testing ML models that are designed for specific tasks \cite{l2017machine}. 
However, the imposition of privacy and data-sharing policies like Global Data Protection Regulations (GDPR) \cite{albrecht2016gdpr} the traditional centralized method of transferring data to the server could present more challenges. In addition to this, other computational challenges are present with this traditional approach \cite{l2017machine}. An alternative solution for creating reliable ML models for tasks needs to be considered.

Federated Learning (FL) is an innovative way to implement ML algorithms and models over decentralized data. First introduced by the research team at Google \cite{mcmahan2017communication}. FL attempts to answer the question \cite{truex2019hybrid} \textit{can we train machine learning models without needing to transfer user data onto a central server?} Since its' introduction, there has been a growing interest in the research community to explore the opportunities and capabilities of FL. The technology enables a more collaborative approach of ML whilst also preserving user privacy by having the data decentralized over the device itself rather than have it over a central server \cite{aledhari2020federated}.

This method of collaborative learning can be explained by using data generated from hospitals as an example. Each hospital generates data from its' smart devices and equipment, using this data could be useful to create an ML model for a specific task. 

However, data from a single hospital may not generate enough data \cite{8970497}. Limitation of data could limit the overall knowledge and performance of the ML model. To create a more robust model that would be able to obtain a higher prediction accuracy access to a larger data set would be better. Other hospitals are also generating their data, however, due to data-sharing policies, it would not be possible to have access to that data. This is where FL and its collaborative nature can thrive where each hospital can train its' own independent ML model that gets uploaded to a central server where all models average out as a global ML model for the specific task. 

This is just one example of the capability of FL is being further researched by a range of industries to maximize this decentralized approach. Industries and applications such as transportation \cite{liu2020privacy}, autonomous vehicles \cite{pokhrel2020federated}, and a range of other Internet of Things (IoT) applications \cite{zhao2019mobile,9424138}. Its implementation can be seen on the mobile application Gboard by Google for predictive texting; the \textit{FederatedAveraging} algorithm to train the ML model over the on-device decentralized data available on mobile devices can improve predictive texting results whilst also reducing the privacy and security risks of the sensitive user data compared to a central server \cite{hard2018federated,chen2019federated,yang2018applied,ramaswamy2019federated}.

\begin{figure}[htp]
	\centering
	\includegraphics[scale=.51]{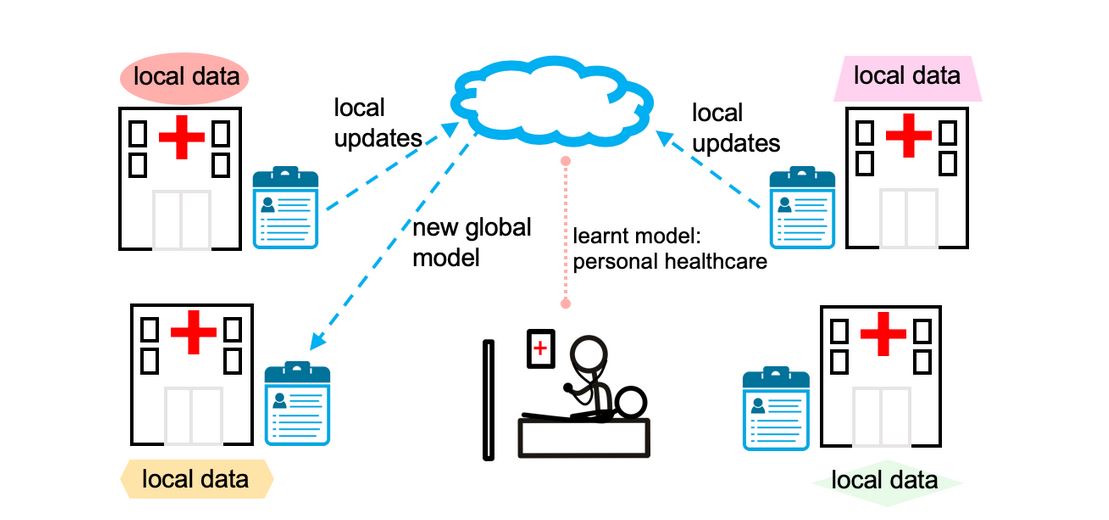}
	\caption{How FL can be used to improve predictive healthcare ML using sensitive data across multiple hospitals \cite{ML_CMU_2019}.}
	\label{image:Hospital_Example_CMU.JPG}
\end{figure}

In addition to preserving data privacy and decentralizing the data, theoretically, the computational power is also split amongst parties that are within the federated network using the decentralized algorithms. Rather than having to rely on the centralized server for training the ML process, the training process can take place on end devices directly.

FL and more of its capabilities are still being discovered, though, as with the introduction of any new technology the more benefits that are reaped of its capabilities the more challenges are exposed. Concerning FL, there are still some privacy, security, communication, and algorithmic challenges that still exist \cite{niknam2020federated}.

In this survey paper, we solely discuss the challenges that are present in \textbf{Communication} concerning FL \cite{kairouz2019advances}. The challenges that arise in an FL setting can come from multiple sources creating a bottleneck. Factors such as end-user network connections that operate at substantially lower rates when compared to network connections that are available at a data center. This form of unreliability could be potentially expensive \cite{kairouz2019advances}. Therefore there has been some research done towards making the Communication factor of an FL environment more efficient. Quantization and Sparsification are two model compression techniques that can be integrated with the FL averaging algorithms to make communication more efficient, along with other techniques that are further discussed in this paper. There has been some research towards this but very limited surveys are found on the topic, therefore, this survey paper aims to present itself by providing a summary of recent research done on the subject so further researchers can benefit from it.
Here:
The remainder of the paper is structured as follows. Section \ref{section:overview} provides an introduction to FL system and how it functions. Section \ref{section:main} introduces and answers the Research Questions (RSQs). Sections \ref{section:discussion} and \ref{section:conclsuion} review the papers where we share our thoughts and navigate towards sharing the future expectations towards this topic of reasearch. 

\section{Problem Statement}
\label{section:overview}
\subsection{Background}
Federated learning was first introduced by the research team at Google \cite{mcmahan2017communication}, the team at Google had the motive to create machine learning models that would be applicable for the wealth of data that is available on mobile devices.
Federated Learning was introduced so users could retain hold on to their data and retain privacy. In FL the ML models can be directly trained on the device itself. 
As the data can be sensitive and also large in quantity in many cases the approach of bringing the model to the data could suit better. This decentralized approach integrated with collaborative learning was given the term Federated Learning.

In this subsection, a general overview and the main components/entities that are available in an FL environment are introduced. Additionally, we will discuss the FL processes from the communication angel.


\subsection{The components of FL systems}
FL system is a learning process where users/participants can collaboratively train the ML model \cite{yang2019federated}. As described by \cite{lim2020federated} there are two main entities or components that are present in an FL environment (i) The data owners, or participants and (ii) The central server. The data owners produce the data on their end devices (e.g. mobile phones), this data generated by the owners is kept private to them and does not leave the device therefore each participant or device has their private dataset. The central server is where the global model stays. It is where the original model is stored and shared across with all the participants that are connected across the FL environment. This model is then a local model for each device i.e. being trained on that independent device's dataset. Once all the device training is complete and model parameters are obtained each device uploads its local model back to the central server. Here at the central is also where each updated local model is uploaded by the participating device and the overall model aggregation takes place by applying aggregation algorithms takes place to generate a new global model.

\subsubsection{The processes of an FL system} 
FL allows a promising approach towards collaborative machine learning whilst preserving privacy \cite{xu2019hybridalpha}. The process of an FL has communication between the two main entities. The two entities, the participating devices and the central server communicate first when devices download a global model from the central server \cite{kang2020reliable}. Each device trains the model based on their private dataset and improves the global model, the computation takes place directly on end devices. Each device then uploads its retrained version of the global model to the central server where an aggregation algorithm merges the model parameters from each participating device into one new generated global model \cite{xu2019hybridalpha}. 

\begin{figure}[htp]
	\centering
	\includegraphics[scale=.51]{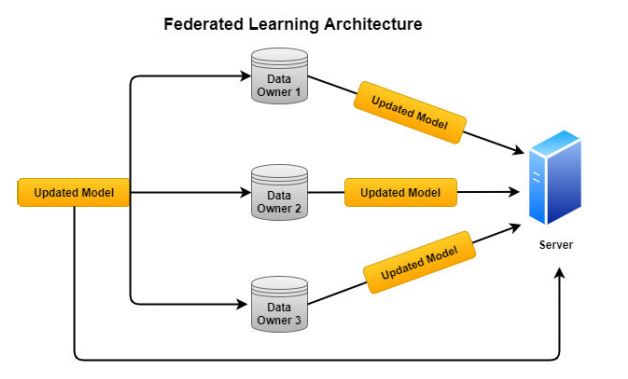}
	\caption{A general overview of how FL works \cite{aledhari2020federated}.}
	\label{image:Federated_Learning_General.JPG}
\end{figure}

\subsubsection{The different types of FL systems}
The base concept of FL is to be able to utilize data that is shared across multiple devices and domains. There are a few ways this data can be distributed across devices, ways such as partitioned by examples or partitioned by features \cite{kairouz2019advances}. The categorization of this data can be a prerequisite step for building an FL environment \cite{mothukuri2021survey}. Some characteristics of data distribution are factors such as heterogeneous data and the clients' participation. In this subsection, a brief introduction is done about the few FL settings that can be applied based on the distribution of data and other characteristics. 

\begin{figure}[htp]
	\centering
	\includegraphics[scale=.51]{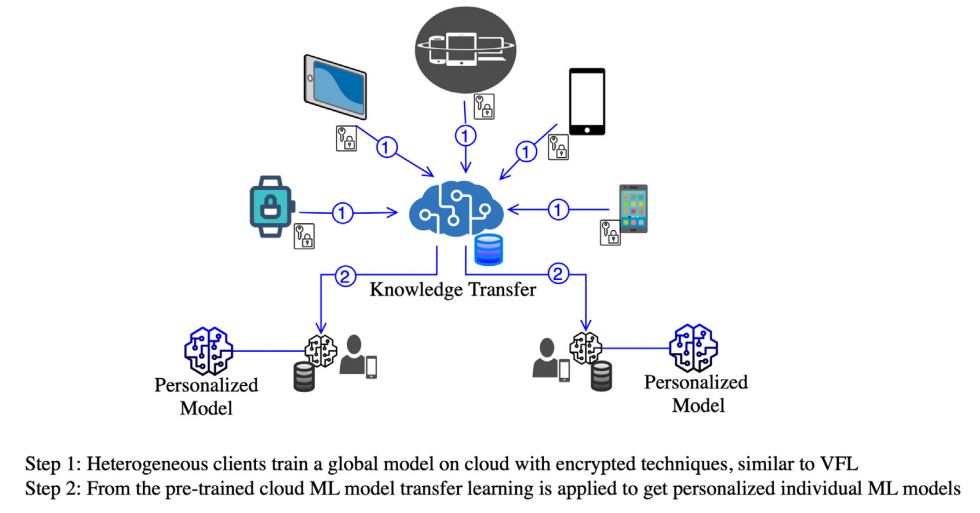}
	\caption{Federated Transfer Learning \cite{mothukuri2021survey}.}
	\label{image:Ftransfer_learning.JPG}
\end{figure}

There are a few types of FL systems, based on the way the data is distributed across the environment \cite{zhu2021federated}, they can be categorized as:
\begin{itemize}
	
	\item \textit{Horizontal Federated Learning:} This type of FL system is when the data from various devices have a similar set of features in terms of the domain but with different instances. This is the original sense of FL learning where data from each domain is homogeneous, see in figure \ref{image:horizontal_fl.JPG}, and contributing together to train the global ML model together \cite{mothukuri2021survey}. This can be explained using the original example that is presented by Google \cite{mcmahan2017communication} wherein the global model is an aggregate of locally trained multiple participating devices \cite{tian2020federboost}.
	
	\begin{figure}[htp]
		\centering
		\includegraphics[scale=.51]{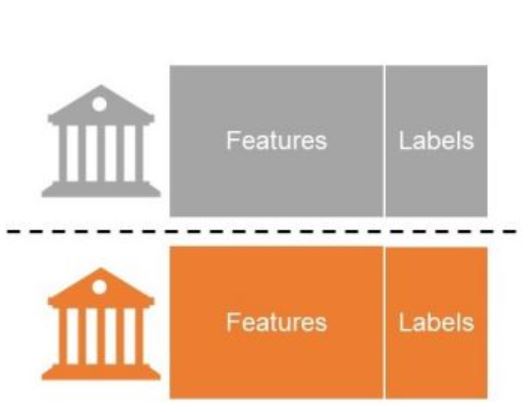}
		\caption{Horizontal FL \cite{tian2020federboost}.}
		\label{image:horizontal_fl.JPG}
	\end{figure}
	
	\item \textit{Vertical Federated Learning:} The data that is distributed across in a Vertical FL setting is data that is common between unrelated domains. This could perhaps be called a feature-based learning scheme as the datasets involved in the training process perhaps share the same sample ID space but may differ in feature space. An example could be where a bank and an e-commerce business in the same city have some form of the mutual user base, shown in figure~\ref{image:vertical_FL.JPG}. The bank has user-sensitive data such as credit card rating, or revenue. Whereas, the e-commerce business has a purchase and browsing history. Here, two different domains can use their data to maybe create a prediction model based on the user and product information \cite{yang2019federated}.
	
	\begin{figure}[htp]
		\centering
		\includegraphics[scale=.51]{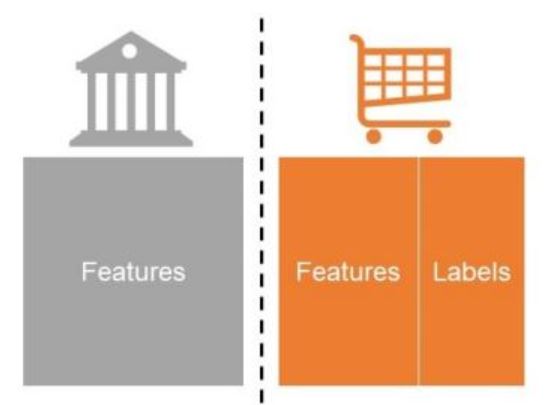}
		\caption{Vertical FL \cite{tian2020federboost}.}
		\label{image:vertical_FL.JPG}
	\end{figure}
	
	\item \textit{Federated Transfer Learning: } This type of system is different from the aforementioned systems where neither the samples nor the features have many similarity \cite{chen2020fedhealth}. An example could be where two data sources such as a bank in the United States and an e-commerce business in Canada are restricted by geography but still have a small intersection with each other being different institutions similar to a vertical FL. However, this is just how the data is partitioned by the ML model is similar to the traditional ML method of transfer learning where the ML model is a pre-trained model on a similar dataset is used. This method can provide better results in some cases compared to a newly built ML model \cite{mothukuri2021survey}, this is further shown in figure \ref{image:Ftransfer_learning.JPG}. 
\end{itemize}

\subsection{Publication Analysis}
Over the recent years, Federated Learning has become a driving research topic for many researchers. Especially those who are interested in decentralized Machine Learning techniques and adhering to the privacy policies that have been imposed. The catalog of research can often times be presented in the form of survey papers. In this subsection we review and introduce the multiple survey papers that have been conducted recently and made available for this domain. Additionally, we highlight our contribution and how our work is different from others.

Authors in \cite{aledhari2020federated} provide a comprehensive and detailed survey on FL. This research done by authors is quite thorough with introducing FL and the different types of FL systems that exist. The authors provide a study with a focus on the software that allows FL to happen, the hardware that platforms, the protocol, and the possible applications for the emerging technology.

Similarly the authors in \cite{xu2021federated} introduce the implications and the practically of FL in healthcare providing general solutions. The authors discuss and introduce some open questions towards data quality, model precision, integrating expert knowledge, etc.
A survey that presents some challenges that can be present in the integration of FL is presented in \cite{lim2020federated}. The authors discuss on Mobile Edge Computing (MEC) and the implementation of FL and how it could act as a dependable technology for the collaborative machine learning approach over edge networks.

The authors in \cite{lyu2020threats} conduct a survey on federated learning and the potential threats that might be available to this approach. As with any new discovery of research and advancements in technology there are always vulnerabilities and potential security threats that can provide a lack of robustness. The survey paper \cite{lyu2020threats} introduces the risk of privacy and how that could potentially be exploited in a FL setting. Furthering, the paper the authors present a brief review on how Poisoning Attacks can interfere and manipulate the outcome of a FL training model, attacks such as data poisoning and model poisoning. Similarly how inference attacks which can also lead towards privacy leakage. Furthering this topic of research, another survey conducted by \cite{mothukuri2021survey} expands more on the threats towards FL as a survey by creating Research Questions with regards to security and privacy and answering them by providing a comprehensive list of threats and attacks, the authors \cite{mothukuri2021survey} also surface some unique security threats exist in the FL environment and what defensive techniques can be implemented towards such vulnerabilities. Defense techniques such as proactive defenses that is a way to guess the threats and the potential risks associated with it. The authors present solutions like Sniper, Knowledge distillation, anomaly detection, and moving target defense.

There are more surveys that are conducted on the topic \cite{li2020federated} \cite{li2019survey} and even though they introduce the topic of communication and the challenges that may be present. There has not been a dedicated research or survey paper that is written towards the challenges that are present in communication and ways to make it efficient. This paper aims to bridge that gap by solely focusing on communication.

\begin{table*}
\centering
\caption{Publication Analysis of survey papers over the past couple of years.}
\resizebox{\linewidth}{!}{%
\begin{tabular}{@{}lllllllll@{}}
\toprule
\multicolumn{1}{l}{\textbf{Reference}} &  
\multicolumn{1}{l}{\textbf{Year}} & 
\multicolumn{1}{l}{\textbf{Objective}} & 
\multicolumn{1}{l}{\textbf{Security}} & 
\multicolumn{1}{l}{\textbf{Privacy}} & 
\multicolumn{1}{l}{\textbf{Communication}} & 
\multicolumn{1}{l}{\textbf{Challenges}} & 
\multicolumn{1}{l}{\textbf{Future Direction}} \\
\midrule[0.1em]

\cite{li2019survey}  &   \multirow{2}{*} {2019} & A comprehensive review of the FL systems, touching and introducing range of FL components.  & \cmark & \cmark & \cmark  & \cmark  &  \cmark  \\

\cite{yang2019federated}  &   & Providing a survey on existing works on FL discussing frameworks, concepts and applications. & \cmark  & \cmark  & \xmark  & \cmark  &  \cmark
\\\cline{1-8}
\cite{lyu2020threats} & \multirow{4}{*} {2020} & Introduction of concept of FL, covering threat model attacks predominantly. & \cmark  & \cmark  & \xmark  & \cmark  &  \cmark \\

\cite{aledhari2020federated} &  & A FL survey focusing on hardware, software and technologies and real-life applications & \cmark  & \cmark  & \cmark  & \cmark  &  \cmark \\

\cite{lim2020federated} &  & Introduction of FL presenting some existing challenges and their solutions & \cmark  & \cmark  & \cmark  & \cmark  &  \cmark \\

\cite{li2020federated} &  & A detailed survey introducing FL and the challenges. & \cmark  & \cmark  & \cmark  & \cmark  &  \cmark
\\\cline{1-8}

\cite{xu2021federated} & \multirow{2}{*} {2021} & A survey of FL in healthcare, covering common topics of introduction of technology, challenges, etc. & \cmark  & \cmark  & \cmark  & \cmark  &  \cmark \\

\cite{mothukuri2021survey} &  & A comprehensive survey posing Research Questions with regards to FL and privacy and security & \cmark  & \cmark  & \cmark  & \cmark  &  \cmark \\
\bottomrule[0.06em] 
\end{tabular}
}
\label{table:analysis}
\end{table*}

A list of recent surveys over the past few years are mentioned in table  \ref{table:analysis}. There is a common theme with most survey papers with survey papers with introducing the technology, presenting the applications and addressing the security and privacy benefits and concerns. In most papers the topic of communication is introduced too, though, it is only a brief surface level part of the paper. This survey paper aims to be the bridge in that gap and present a survey paper that focuses solely on the communication component for FL.

\section{Research Questions \& Communication Efficient methods}
\label{section:main}

Communication between the participating devices and the central server is an essential step for an FL environment. The model is communicated over rounds of communication that download, upload and train the ML model. However, as mentioned, this comes with its' challenges. To better gain more insight into the challenges that are present in an FL environment with regards to communication we pose two Research Questions (RSQs). The first \textbf{RSQ1} aims to provide a brief analysis of \textit{What are some of the challenges that are presented in FL with regards to communication?} The second \textbf{RSQ2} provides an analysis in-depth about various methods and approaches that can be implemented to answer \textit{How can we make communication more efficient in an FL environment?}

\subsection{RSQ1 - What are some of the challenges that are presented in FL with regards to communication?}
As aforementioned in Section \ref{section:overview} about the general working of the FL environment the central server shares the global model across the network to all the devices that are participating. The total number of devices that are participating in this FL environment can sometimes be in millions and the bandwidth over which the devices are connected to the environment could be relatively slow or unstable \cite{konevcny2016federated}. In an FL training environment, there can be many rounds of communication that exist between the central server and all the participating devices. Over a singular communication round the global model is shared across all the devices in the FL environment and each participating device downloads the global model to train it on their local dataset. A version of that is uploaded back to the environment to the central server. Therefore there is a constant downlink and uplink during the communication rounds \cite{zheng2020design}. Though due to limited bandwidth, energy and power on the device end these rounds of communication can be slow \cite{li2020federated}. Even with these challenges, the overall communication cost of sharing model updates is relatively lower than sharing copious amounts of data from the devices to a central server; it is still important to preserve the communication bandwidth further to make it more efficient \cite{mothukuri2021survey}. In this subsection, we introduce the overheads \cite{luping2019cmfl} or challenges that could create a communication bottleneck in an FL environment.
\newline

\begin{itemize}
	\item \textit{Number of Participating Devices:} Though having a high number of participating devices in an FL environment has its' advantage wherein the ML model could be trained on more data and a possible increase in performance and accuracy. However, the large number of devices that are participating in multiple FL training rounds at the same time could create a communication bottleneck too. In some cases, a high number of clients could lead to an increase in the overall computational cost too \cite{chen2019communication,mothukuri2021survey}. 
	
	\item \textit{Network Bandwidth:} Though in contrast to the traditional ML approach, the FL approach does reduce the cost substantially, however, the communication bandwidth still needs to be preserved \cite{mothukuri2021survey}.  The participating devices may not always have the bandwidth needed and could be participating under unreliable network conditions. Factors such as having a difference between upload speed and download speed that could result in delays such model uploads by participants \cite{lim2020federated} to the central server which could lead to potential bottleneck leading to disrupting the FL environment. 
	
	\item \textit{Limited Edge Node Computation:} The computation is now dependant on edge devices rather than powerful GPUs and CPUs. The edge devices could have limitations towards computation, power resources, storage and as aforementioned limited link bandwidth \cite{shi2020communication}. The authors in \cite{shi2020communication} give a comparison in training time between a central server and an edge device. They elaborate that an Image Classification model with over 60 million parameters can be trained in just a few minutes over a GPU reaching speeds of 56Gbps. However, even with a powerful smartphone connected over 5G it could take much longer reaching an average speed of 50Mbps.
	
	\item \textit{Statistical Heterogeneity:} Another possible source for a communication bottleneck or where communication costs can rise could be statistical heterogeneity where the data is non independent and identically (non-i.i.d.) distributed \cite{li2018federated}. In a FL environment, the data is only locally present on each participating device. It is gathered and collected by the participants on their independent devices based on their usage pattern and local environment. An individual participant's dataset in an FL environment could not be representative of the population distribution of the other participants and their datasets \cite{mcmahan2017communication}. The size of data gathered and distributed amongst devices can typically vary heavily \cite{sattler2019robust}. Therefore, this type of fluctuation in the size of the dataset could affect communication by causing a delay in model updates and other attributes. A device with a larger dataset could take longer to update, whereas a device with a smaller dataset could be done with updates. However, the global model might not be aggregated until all individual client models are trained and uploaded causing a bottleneck.
	\newline
	
\end{itemize}

The aforementioned constraints listed are just some of the discovered challenges that could be possible sources towards creating a communication bottleneck.

\subsection{RSQ2 - How can we make communication more efficient in an FL environment?}

To train the global ML model with decentralized data, the global model needs to be downloaded on the participating devices on that federated network. This allows the data generated by those remote devices to remain preserved on the device whilst subsequently also making it possible to improve the ML model. The steps that make this possible can be summarized into three communication steps (1) the global model needs to be shared across devices within the federated network, these can sometimes be millions of IoT devices of mobiles phone (2) The model is then downloaded by the devices and trained locally on-device on the private dataset that is generated by those devices (3) the ML model is uploaded back to the central server where it is pooled with numerous other models that have been uploaded to aggregate them all together and find a federated average to generate a new and updated global model. Considering the steps of communication that part-take to obtain an improvement in the global ML model it is important to seek out the most communication-efficient methods that could make the transfer of data from (1) to (2) and onto (3). 

In this subsection, we present the recent findings and efforts that are made by the research community to improve the communication of the ML model over the federated network. Findings that are tackling constraints where devices can drop out due to poor or limited network bandwidth. In addition to this, other constraints too where data is sampled differently and how these could affect communication.
We list some research towards discovering different methods that could help with making communication more efficient. Methods such as Local Updating, Client Selection, Reduction in Model Updates, Decentralized Training \& Peer-to-Peer Learning, and various Compression Schemes.

\begin{figure}[htp]
	\centering
	\includegraphics[scale=.61]{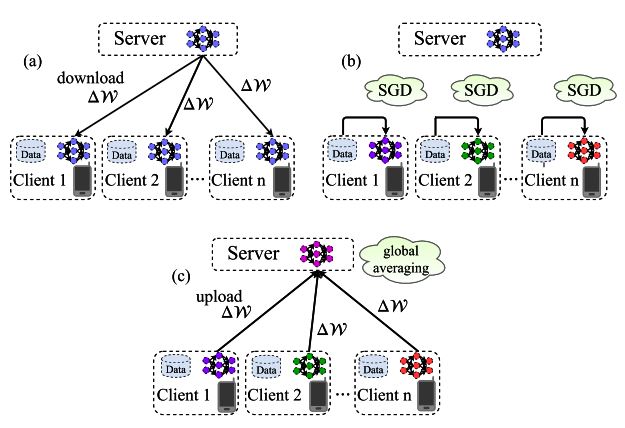}
	\caption{\cite{sattler2019robust} illustrate a complete round of distributed Stochastic Gradient Descent (SGD) model. In (a) clients on the federated network synchronize with the server and download the global ML model. In (b) the global model is trained by each client on their local data, adjusting the weighted average of the ML model as per. In (c) the new average achieved by each client is updated onto the server where each update is used to obtain a new weighted average for the global model.}
	\label{image:communication_rounds.JPG}
\end{figure}



\subsubsection{Local Updating}
Distributed ML that exists in data centers have become popular with integrating mini-batch optimization methods, however, there are still some constraints and limitation on flexibility. This form of batch optimization has constraints over a more federated setting that includes both communication and computation challenges \cite{zhang2015deep,li2020federated}. Though the overall objective of local updating is mostly focusing on the aforementioned point (2) i.e. to fit and train the ML model locally on-device using the data that is generated by those devices, however, having to compute locally over a singular communication round and then applying those updates to the central server is not that efficient due unexpected dropouts of devices or synchronization latency due to poor network connectivity \cite{zhao2018federated}. Authors in \cite{mcmahan2017communication} also highlight that it is very important to have a form of flexibility when considering optimization method. Flexibility towards client participation and local updates. There are a few ways towards achieving this flexibility and in turn improving the overall communication efficiency. Techniques such as primal-dual methods can offer a form of flexibility where local devices can use local parameters to train the global model locally to find arbitrary approximations based \cite{smith2017cocoa}. This leveraging and breaking down the overall objective of the global model can allow for problems or training to be solved in parallel over each communicated round \cite{zhang2015deep}. In this subsection, we review some of the methods that have come surface with regards to local updating techniques that can make communication more efficient.

In an FL setting the data can be distributed unevenly across devices and may not always be identical. This can be considered that the distributed data across the FL environment is in a non-independent and identically distributed manner i.e. non-iid. Having the datasets in a non-id state could challenge the testing accuracy of an FL model \cite{zhao2018federated}. The testing accuracy of a locally trained ML model is important as it will contribute to the global model. Therefore having an accuracy of a local model that performs poorly could also deteriorate the overall performance of the global model.

In \cite{briggs2020federated} the authors introduce their way of combining FL with Hierarchical Clustering (HC) techniques to reduce the overall communication rounds within the FL protocol. The clustering step introduced can cluster and separate the clients by the similarity of the local updates to the global model. These similarities are the weights that are achieved by locally updating the local model on each device. Upon testing and comparing their integrated FL + HC technique the authors concluded the communication rounds are reduced as per a comparative on the Manhattan distance metric \cite{singh2013k,briggs2020federated}.

Due to the large number of connected devices that are attempting to communicate their locally updated model parameters with the central server, (2), this could cause a communication bottleneck due to bandwidth in some instances
\cite{reisizadeh2020fedpaq}. The authors in \cite{reisizadeh2020fedpaq} introduce a new way to counter this problem, their FedPAQ framework i.e. Federated Periodic Averaging and Quantization framework tackles this problem with its' features. Its feature of Periodic Averaging feature, in particular, helps reduce the number of communication rounds. In contrast to other training methods where each participating device sends their ML models to synchronize through the parameter servers over each iteration resulting in increasing the number of communication rounds between the devices and central sever. However, this method of Periodic Averaging method can allow a solution using the Stochastic Gradient Descent (SGD). The parameters and local updates of each device can be synchronized with the central server where a periodic average of models takes place.
This is done by adjusting the parameter that corresponds to the number of iterations that occur locally on the device itself. Other features such as Partial Node Participation and Quantized Message-Passing of the FedPAQ also reduce communication overhead.
The Quantized Message-Passing is further discussed in the Compression Schemes subsection.

The aforementioned FedAvg algorithm is used for its simplicity and in turn, is about to reduce communication costs in an FL environment. The algorithm tackles the challenges presented over communication by performing numerous updates on available devices before communicating with the central server. However, in some cases such as the authors in \cite{karimireddy2020scaffold} state that over heterogeneous data the FedAVG algorithm could introduce 'drift' towards the updates of each client or device that is participating. This 'client drift' if will result in slow and unstable convergence and could persist if all clients participate in training rounds with full batch gradients. 
The authors run their analysis and determine that the FedAvg algorithm full batch gradients and matching lower bounds over no client sampling to conclude that it is slower than the Stochastic Gradient Descent (SGD) over some parameters \cite{karimireddy2020scaffold}. To over this challenge the authors created their framework called Stochastic Controlled Averaging (SCAFFOLD) algorithm that can fix the drift \cite{karimireddy2020scaffold}. For their analysis, the SCAFFOLD algorithm performs well and provides reliable convergence rates as SGD even for non-iid data. It also takes the advantage of the similarity that exists within the clients and reduce communication overload \cite{shamir2014communication,reddi2016aide}

An approach introduced by authors in  the \cite{li2019feddane} called FedDANE tackles some of the practical constraints that are present with FL. The approach is a culmination of methods introduced in \cite{reddi2016aide, shamir2014communication}. FedDANE collects gradient updates from subset of devices during each of the communication rounds. FedDANE works well with low client participation settings too.


Local updating is an essential part of Federated Learning. Each device needs to compute local updates so a better overall global model can be generated and all participating devices can benefit from it. Making the process of local updating more efficient could indeed make communication more efficient.

\subsubsection{Client Selection}
Client Selection is an approach that can be implemented to make communication more efficient by reducing the costs by restricting the number of participating devices so only a fraction of the parameters is updated over the communication round \cite{xu2021federated}. In this subsection, we introduce some of the research that is done with regards to this and how beneficial it could be towards easing communication overload.

An approach introduced by authors \cite{abdulrahman2020fedmccs} tackles the limitations that are present in communication over an FL setting. The authors \cite{abdulrahman2020fedmccs} create a multi-criteria client selection model for IoT devices over an FL setting. Their FedMCCS approach considers all the specifications and network conditions of the IoT device for client selection. Things such as CPU, Memory, Energy and Time are all considered for the client resources to determine whether the client would be able to participate in the FL task. The FedMCCS approach considers more than one client and over each round, the number of clients participating is increased. From their analysis FedMCCS in comparison to other approaches can outperform by reducing the total number of communication rounds to achieve reasonable accuracy.

Another factor that is a vital property of FL according to authors in \cite{xu2020client} is the varying significance of learning rounds. This comes to the surface when the authors realize that the learning rounds are temporally interdependent but also have varying significance towards achieving the final learning outcome. This conclusion comes from running numerous data-driven experiments, the authors create an algorithm that utilizes the wireless channel information but can achieve long-term performance guarantee; the algorithm results in providing a desired client selection pattern that is adapted to network environments. 

Authors in \cite{nishio2019client} introduced a framework that specifically tackles the challenges when it comes to client selection. The authors called their framework FedCS. Their goal is that when it comes to Client Selection in a standard FL setting it can sometimes be a random selection of clients (or devices), however, with their approach of Client Selection they break into two steps. First, Resource Request, where client information such as state of the wireless channel and computational capacities, etc are requested and shared with the central server. Second, to estimate the time required Distribution and Scheduled Update and Upload steps are taken. With their framework, the overall approach of the Client Selection is to allow the server to aggregate as many clients within a certain time frame. Knowing these factors such as data, energy, and computational resources i.e. used by the devices could better meet the requirements of a training task and possibly affect energy consumption and bandwidth cost. 

To better assess whether a client can sufficiently participate or not a resource management algorithm introduced by authors in \cite{anh2019efficient} adopts a deep-Q learning algorithm that could allow the servers to learn and make optimal decisions without having to know prior network knowledge. Their Mobile Crowd Machine Learning (MCML) algorithm addresses the constraints that are present in mobile devices reducing the energy consumed along with making the training and communication time more efficient too.

Authors run their analysis providing a method that practices biased client selection strategies \cite{cho2020client}. As per their analysis, the selection of bias can affect the convergence speed, the bias of their client selection works towards clients that have a higher local loss and are achieving faster error convergence. With this knowledge, the authors a framework that is a communication and computation efficient framework. Calling it Power-Of-Choice. Their framework has the agility to trade-off between the bias and the convergence speed. The results the authors achieve are computed relatively faster and provide higher accuracy of results than the baseline random selection methods \cite{cho2020client}.

Uplink and downlink communication with participating clients in a federated network is necessary. Having to communicate with clients that have dependable network bandwidth and energy sources could aid towards achieve a well-trained global model more efficiently. Client selection techniques implemented can aid in reducing the cost of overall communication that is needed to achieve a dependable global model.

\subsubsection{Reducing Model Updates}
Once the global model is downloaded by connected devices in the FL environment each device starts to train the devices locally. As each device computes and updates the model these updates are communicated back to the central server \cite{mcmahan2017communication}. The number of communication rounds between the devices and central server can be costly and perhaps having fewer but more efficient model updates could be a solution. In this section, we introduce some techniques that discuss a possible reduction in communication for these model updates and potentially reduce the cost. 

Authors in \cite{kamp2018efficient} introduce an efficient way of training models, their proposed approach adapts to the concept of drifts trains models equally well through different phases of model training. Their approach leads to reducing the communication substantially without depreciating the predictive performance of the model.

In another study \cite{bui2018partitioned}, authors introduce a Partitioned Variational Inference (PVI) for probabilistic models that work well over federated data. They train a Bayesian Neural Network (BNN) over an FL environment that is allowed for both synchronous or asynchronous model updates across many machines, their proposed approach along with the integration of other methods could allow a more communication-efficient training of BNN on non-iid federated data. 

In contrast to most of the current FL methods that include iterative optimization techniques over numerous communication rounds \cite{smith2017federated,mcmahan2017communication}, authors in \cite{guha2019one} introduce a one-shot communication round approach where only a single round of communication is done between the central server and the number of connected devices \cite{guha2019one}. The authors do suggest as opposed to computing increments, each device trains a local model to completion and then applies ensemble methods to effectively capture information regarding device-specific models. Applying ensemble learning techniques could be better suited for global modelling than averaging techniques \cite{guha2019one}.

In an FL setting the rate at which model convergence occurs can sometimes take a large number of communication rounds creating a delay towards model training whilst simultaneously increasing network resources \cite{nguyen2020fast}. An intelligent algorithm called FOLB is introduced by authors in \cite{nguyen2020fast}. The algorithm performs smart sampling of participating devices over each round to optimize the expected convergence speed. The algorithm can estimate the participating device's capabilities, this is done by adapting to devices aggregations.

If the mere frequency in which the model updates are shared is reduced that means the overall communication between the devices and the central server is reduced too. Having an effective way of determining how these updates are computed and reducing the overall communication rounds that are needed. Fewer rounds could result in more efficiency.

\subsubsection{Decentralized Training \& Peer-to-Peer Learning}
FL in a way is like practicing ML in a decentralized manner. However, FL does allow a more peer-to-peer learning approach wherein each node that is trained can benefit from the other node that is trained on the FL network too. Even in decentralized training similar challenges of communication exist and to tackle it different methods like compression \cite{tang2018communication} can be used. In this subsection how decentralized or peer-to-peer learning is utilized or integrated into an FL environment.

\begin{figure}[htp]
	\centering
	\includegraphics[scale=.61]{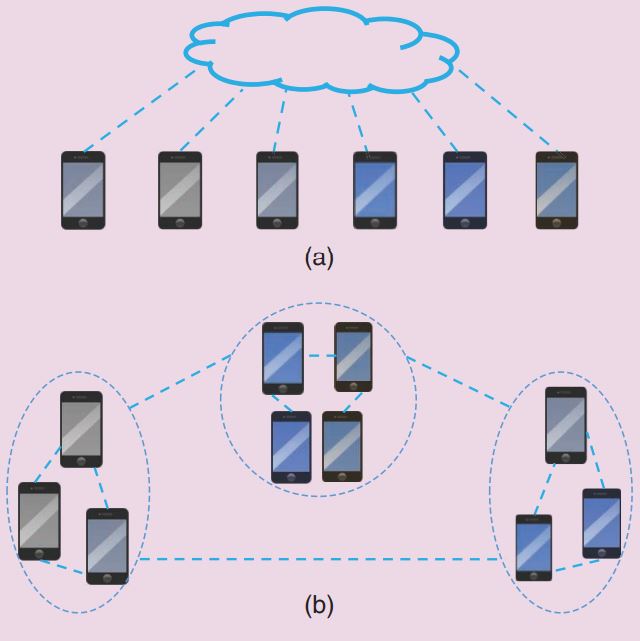}
	\caption{An overview of star-network topology. \cite{li2020federated}}
	\label{image:distributed_training.JPG}
\end{figure}

In an FL environment, as aforementioned, there is a central server that has the original model. The central server is where all the devices that are connected to the network update their model too. Essentially, all devices that are participating in the FL environment are connected to the central server. As stated by \cite{li2020federated} the star network is the predominant communication topology in an FL setting shown in Figure \ref{image:distributed_training.JPG}. 
However, there is some research on whether a different communication topology where the participating devices only communicate with one another. A Peer-to-Peer learning experience or decentralized training network, and whether this would be a more efficient way to communicate in an FL environment. 
Authors in \cite{li2020federated} state that in traditional data center environments decentralized training can appear to be faster than centralized training especially when constraints of low bandwidth and high latency are faced when operating on networks. Though this is not to say that decentralized training does not have its' constraints authors in \cite{reisizadeh2019robust} state that nodes computation time on nodes can slow down the convergence of a decentralized algorithm, in addition to this, the sometimes, large communication overhead could inalso further mitigate this. To overcome these, the authors in \cite{reisizadeh2019robust} proposed QuanTimed-DSGD algorithm, a decentralized and gradient-based optimization, that imposes iteration deadlines for nodes and nodes exchange their quantized version of the models. 

In the FL environment, peer-to-peer learning could be where each device only communicates with its neighbours and updates. The participating devices or clients updates their model on their dataset, and aggregates it along with the model updates from their neighbours \cite{xu2020federated}. Furthering that, authors in \cite{he2019central} build a framework for an FL environment towards a generic social network scenario. Their Online Push-Sum (OPS) method handles the complex topology whilst simultaneously having optimal convergence rates. 

Similarly, \cite{lalitha2019peer} propose a distributed learning algorithm where there is no central server but instead, the participating devices practice peer-to-peer learning algorithm to iterate and aggregate model information with their neighbour to collaboratively estimate the global model parameters. The authors base an assumption suggesting that FL setting wherein a central server exists and communicates the global model can incur large communication costs. In their approach, the devices are already distributed over the network where communication occurs only with their one-hop neighbours.

The authors in \cite{roy2019braintorrent} that further provides another avenue for peer-to-peer learning and not depending on the central server as a single trusting authority. The authors \cite{roy2019braintorrent} create their framework BrainTorrent that does not rely on a central server for training, their proposed peer-to-peer framework can is designed to motivate a collaborative environment. According to their analysis, the absence of a central server makes the environment resistant to failure but also precludes the need for a governing central body that every participant trusts. 

Theoretically, the application of decentralized training in an FL setting could reduce the communication cost when compared with a central server \cite{li2020federated} though there more research done on this and there are further avenues it can take too. 

\begin{table*}[!ht]
\renewcommand{\arraystretch}{1.2}
\caption{\label{tab:table-name}Research done towards reducing the overall Communication costs and overheads in a FL.}
\centering
\resizebox{\linewidth}{!}{
\begin{tabular}{lp{2.5cm}p{4cm}p{12cm}}
\hline
  \multicolumn{1}{l}{\textbf{Reference}}&
  \multicolumn{1}{l}{\textbf{Section}}&
  \multicolumn{1}{l}{\textbf{Model and Technology}}&
    \multicolumn{1}{l}{\textbf{Remarks}}
  \\\hline
\cite{briggs2020federated}  &
  \multirow{3}{*}
  {Local Updating}
 & Hierarchical Clustering Technique & A FL+HC technique that separates client clusters similarity of local updates \\\cline{3-4}\cline{1-1}
\cite{reisizadeh2020fedpaq} &     & FedPAQ & Using periodic averaging to aggregate and achieve global model updates
\\\cline{3-4}\cline{1-1}
\cite{karimireddy2020scaffold} &     & SCAFFOLD Algorithm & An algorithm that provides better convergence rates over non-iid data
\\\cline{3-4}\cline{1-1}
\hline
\cite{sattler2019robust} &
  \multirow{5}{2.5cm}  
  {Compression Schemes - Sparsification}
 & STC method & Providing compression for both upstream and downstream communications  \\\cline{3-4} \cline{1-1}
\cite{rothchild2020fetchsgd} &     & FetchSGD & Compresses the gradient based off of client's local data \\\cline{3-4}\cline{1-1}
\cite{li2020ggs} &     & General Gradient Sparsification (GSS)  & The batch normalization layer with local gradients achieved from GSS could mitigate the impact of delayed gradients and not increase the communication overhead  \\\cline{3-4}\cline{1-1}
\cite{hu2020sparsified} &     & CPFed & A sparsified masking model that can provide compression and differential privacy \\\cline{3-4}\cline{1-1}
\cite{sattler2019sparse} &     & Sparse Binary Compression (SBS) & Introducing temporal sparsity where gradients are not communicated after every local iteration  \\\cline{3-4}\cline{1-1}
\hline
\cite{reisizadeh2020fedpaq} &
  \multirow{5}{2.5cm}  
  {Compression Schemes - Quantization}
 & FedPAQ & Using quantization techniques based upon model accuracy \\\cline{3-4} \cline{1-1}
\cite{amiri2020federated} &     & Lossy FL algorithm (LFL) & Quantizing models before broadcasting  \\\cline{3-4}\cline{1-1}
\cite{dai2019hyper} &     & Hyper-Sphere Quantization (HSQ) framework & Ability to reduce the cost of communication per iteration  \\\cline{3-4}\cline{1-1}
\cite{shlezinger2020federated} &     & UVeQFed & With the algorithm convergence of model minimizes the loss function \\\cline{3-4}\cline{1-1}
\cite{liu2021hierarchical} &     & Heir-Local-QSGD & A technique that naturally leverages client-edge-cloud network hierarchy and quantized models updates   \\\cline{3-4}\cline{1-1}
\hline
\cite{roy2019braintorrent} &
  \multirow{2}{2.5cm}
  {Decentralized Training or Peer-to-peer learning}
 & BrainTorrent & A peer-to-peer learning framework where models converge faster and reach good accuracy \\\cline{3-4}\cline{1-1}
\cite{reisizadeh2019robust} &    &   QuanTimed-DSGD  & decentralized gradient-based optimization imposing iteration deadlines for devices \\\cline{3-4}\cline{1-1}
\hline
\cite{abdulrahman2020fedmccs} &
  \multirow{4}{*}
  {Client Selection}
 & FedMCCS & A multi-criteria client selection that considers that considers IoT device specification and network condition \\\cline{3-4}\cline{1-1}
\cite{xu2020client} &    & Resource Allocation Model & Optimizing learning performance on how clients are selected and how bandwidth is allocated \\\cline{3-4}\cline{1-1}
\cite{nishio2019client} &    & FedCS & The framework allows the server to aggregate as many clients within a certain time-frame \\\cline{3-4}\cline{1-1}
\cite{cho2020client} &    & Power-Of-Choice  & A communication and computation efficient client selection framework \\\cline{3-4}\cline{1-1}
\hline
\cite{kamp2018efficient} &
  \multirow{3}{2.5cm}
  {Reduced Model Updates}
 & A decentralized deep learning model & Ability to handle different phases of the model training well \\\cline{3-4}\cline{1-1}
\cite{bui2018partitioned} &    &  a Partitioned Variational Inference (PVI)   & A Bayesian Neural Network over FL that is synchronous and asynchronous for model updates across machines \\\cline{3-4}\cline{1-1}
\cite{guha2019one} &    & One-shot federated learning & A single round of communication done between central server and connected devices  \\\cline{3-4}\cline{1-1}
\cite{nguyen2020fast} &    & FOLB & Intelligent sampling of devices in each round of model training to optimize the convergence speed   \\\cline{3-4}\cline{1-1}
\hline
\end{tabular}
}
\end{table*}

\subsubsection{Compression Schemes}
In addition to the local updating methods, some ML compression techniques may also be implemented towards reducing the total number of rounds between the centralized server and the devices that are connected on the federated network. Compression schemes like quantization and sparsification that are sometimes integrated with the aggregation algorithms can sometimes provide reasonable training accuracy results whilst at the same time also reducing the overall communication cost. The authors in \cite{kairouz2019advances} list the objectives set for compression schemes and federated learning as such:
\begin{enumerate}[label={\alph*.}]
	\item Reducing the size of the object / ML model from the clients to the server i.e. used to update the overall global model 
	\item Reducing the size of the global model that shared with the clients on the network, the model on which the clients start local training using the available data
	\item Any changes that are made to the overall training algorithm that makes training the global training model more computationally efficient
\end{enumerate}

From the list of objectives \textbf{A} could have the highest effect on the overall running time of around therefore in reducing that would directly result in reducing the overall communication cost; the clients on a federated network generally have a slower upload bandwidth compared to a download. Therefore compressing the ML models and potentially reducing the uplink/downlink exchanges could result in reducing the communication cost \cite{xu2020federated}.  

In this subsection, we review the research that is done by the community towards compression schemes to make communication more efficient in a federated network. We begin by sharing the research that is done towards compression schemes, then we further divide the research concerning methods of sparsification and quantization.

\textbf{\textit{Sparsification:}} Sparsification techniques is a type of communication technique that can be implemented in an FL setting to compress the model when they are being communicated across the server, in this section, we review the research that has been conducted towards making integrating sparsification in an FL environment and making communication more efficient. 

The authors in \cite{sattler2019robust} present a sparse ternary compression (STC) method that not only adheres to the compression requirements of an FL setting and environment but also provides compression for both upstream and downstream communications. They run an analysis over FL models over various datasets and architectures, in their analysis, they conclude that some factors of an FL environment are hugely dependant on the convergence rate of the averaging algorithm. From their analysis, the authors deduce that factors where training is done on non-iid small portions of data or when only a subset of clients participate in communication rounds can reduce the convergence rate. 
However, the proposed model of STC is a protocol that compresses communication via sparsification, ternanization, error accumulation and the optimal Golomb encoding. The robust technique provided by authors \cite{sattler2019robust} converges relatively faster when compared to other averaging algorithms like FedAVG over both factors of non-iid data and a lesser number of iterations that are communicated. The proposed method is also highly effective when the communication bandwidth is constrained.

Sparse client participation is another challenge that needs to be overcome in an FL environment, authors introduce a FetchSGD algorithm that can help towards achieving this \cite{rothchild2020fetchsgd}. FetchSGD overall aids with communication constraints of an FL environment by compressing the gradient that is based on the client's local data. The data structure Count Sketch \cite{spring2019compressing} is used to compress the gradient before it is uploaded to the central server. The Count Sketch is also used for error accumulation. According to authors in \cite{li2020ggs}, one key problem with regards to federated setting and communication is the communication overhead that is involved in parameters synchronization. The authors inform that this overhead wastes bandwidth and increases training time whilst impacting the overall model accuracy. To tackle this, the authors propose a General Gradient Sparsification (GGS) framework for adaptive optimizers. The framework consists of two important mechanisms that are batch normalization updates with local gradients (BN-LG), and gradient correction. The authors determine that updating the batch normalization layer with local gradients could mitigate the impact of delayed gradients and not increase the communication overhead. The authors run their analysis over several models such as AlexNet, DenseNet-121, CifarNet, etc, and achieve high accuracy results concluding gradient sparsification does have a significant impact in reducing the communication overhead. 
The authors in \cite{hu2020sparsified} introduce a compression scheme that is both a communication efficient and a deferentially private federated learning scheme. The authors call it CPFed. The challenge the authors face when addressing both issues together is that one that rises from data compression. Techniques used for compression could sometimes lead to an increased number of training iterations required for achieving some desired training loss due to the compression errors, however, differential privacy could deteriorate with regards to the training iterations. To overcome this paradigm, their proposed CPFed is based on a sparsified privacy masking technique that adds random noise to model updates along with an unbiased random sparsifier before updating the model. The authors can achieve high communication efficiency through their proposed model. 

Authors in \cite{sattler2019sparse} propose a compression technique that could drastically reduce the communication cost for a distributed training environment. The framework introduced by the authors is a Sparse Binary Compression (SBC) technique, their method integrated techniques that are already present in communication delay and gradient sparsification with a novel binarization method. According to their research, the authors find the current gradient information for training neural networks with SGD is redundant. Instead, the authors utilize communication delay methods introduced in \cite{mcmahan2017communication} to introduce temporal sparsity where gradients are not communicated after every local iteration. From their findings, the authors conclude that in a distributed SGD setting both communication delay and gradient sparsity can both be treated as independent types of sparsity techniques. These methods provide higher compression gains, though, in their findings the authors also find a slight decrease in Accuracy. 

\textbf{\textit{Quantization:}}
The communication between the devices and the central server includes sharing model updates and parameters that have occurred on the device end. This update of parameters can be strenuous when it comes to up-linking the model. To aid with this, another compression method referred to as quantization can bring the model parameters to a reasonable size without compromising much on model accuracy \cite{quantmedium}. In this subsection, we introduce different techniques and frameworks that have integrated quantization towards making communication more efficient for an FL setting.

The authors in \cite{reisizadeh2020fedpaq} utilize their framework FedPAQ to reduce the overall communication rounds and the bearing cost. A feature of their framework is to Quantize Message-Passing. Communication bandwidth will mostly be limited in an FL setting, with limited uplink bandwidth on participating device end could increase the communication cost making it more expensive.
Authors \cite{reisizadeh2020fedpaq} employ quantization techniques \cite{jiang2018linear} on the up-links; every local model is quantized before being uploaded hence reduce the overall communication overhead.

Furthering and employing quantization techniques authors in \cite{amiri2020federated} use them for their FL analysis. In contrast to typical FL models where the global model from a central server is downloaded by all devices and then subsequently updated and so and so forth. The authors from their analysis discovered that applying quantization techniques to the global model can help towards making communication more efficient. The Lossy FL algorithm (LFL) created by the authors quantizes the global model before it broadcasts and shares it across with the devices. The local updates that take place on the device are uploaded on the global central server are also quantized. The FL environment is hugely dependant on bandwidth and the authors \cite{amiri2020federated} for their analysis study how well the quantized compressed global model performs to provide an estimate for the new global model from the local updates of the devices. They compare their results with another analysis of a lossless standard approach of FL to conclude that their LFL and the technique of quantizing global model updates provide a significant reduction in communication load. 

Traditionally, compression algorithms are trained for a setting where there is a high-speed network, places such as data centers \cite{koloskova2019decentralized} but in an FL setting these algorithms might not be directly very effective. To further tackle the communication bottleneck that is created due to the network and the other aforementioned reasons another quantization compression technique is proposed by \cite{dai2019hyper}. To achieve a trade-off between communication efficiency and accuracy authors in \cite{dai2019hyper} propose a Hyper-Sphere Quantization (HSQ) framework. HSQ can reduce communication cost low per-iteration which is ideal for an FL environment, the framework utilizes vector quantization techniques that shows an effective approach towards achieving gradient compression and simultaneously not comprising over on convergence accuracy. 

In another study, the authors \cite{shlezinger2020federated} suggest that the communication channel and transfer of model parameters between the users to the central server has a throughput that can be typically constrained. The authors whilst doing their research encountered that alternative methods to aid with this could provide dominant distortion of results. This lead \cite{shlezinger2020federated} to create a quantization method that is more efficient towards facilitating the model transfer in an FL setting. Utilizing quantization theory methods, the authors design quantizers that are suitable for distributed deep network training. Understanding the requirements that are needed for quantization FL setting, the authors can propose an encoding-decoding strategy. Their proposed scheme shows potential for an FL setting as it performs relatively well compared to previously proposed methods. Furthering their research towards quantization theories the authors in their review approach it from identifying unique characteristics regarding the conveyed trained models over rate-constrained channels \cite{shlezinger2020uveqfed}. The authors propose a Universal Vector Quantization technique for FL; a quantization technique that would be suitable for such settings calling it UVeQFed. For their research, the authors demonstrate that combining universal vector quantization methods can yield a system where compression of trained models induces only a minimum distortion. Analyzing the distortion further the authors determine that the distortion reduces substantially as the number of users grows.  

Authors in \cite{liu2021hierarchical} introduce a Hierarchical Quantized Federated Learning technique that can leverage client-edge-cloud network hierarchy and quantized model updates. Their Heir-Local-QSGD algorithm performs partial edge aggregation and quantization on the model updates that can result in improving the communication efficiency in an FL environment.

\section{Discussion}
\label{section:discussion}
This paper provides an introduction towards FL whilst retaining focus on the communication component of FL. Communication is quite an essential part of the Federated Learning environment. It is essentially how the global machine learning model is transmitted from the central server to all the participating devices. Similarly, the model is then trained on the local data that is available on the devices, and then uploaded it back to the central server. This constant communication requires a lot of download and upload using a reliable network bandwidth. As it is not always the case, limited bandwidth or poor client participation this can create a communication lag in completing the FL training. Techniques to make the rounds of communication more efficient are shared in this paper. For example a wide range of compression scheme methods mentioned in this paper can help towards reducing the communication overheads and reducing the costs to some factor. Even though the methods presented in this paper do make the communication front more efficient they are all implemented using a range of resources such as different computational power and datasets. There are dependable ways towards which communication rounds can be addressed, however, due to the range of solutions presented it could be concluded that a culmination of techniques implemented together could result in the most communication robust solution for a FL setting.

\section{Conclusion \& Future Expectations}
\label{section:conclsuion}
This survey paper aimed to provide a bridge between the gap that was present on the topic of communication in FL. This paper presented the challenges and constraints that are present concerning this. Such as bandwidth and limited computation. We posed these challenges in the form of two RSQs that both introduced the problems and provided a list of solutions that have been applied towards making communication more efficient. We believe that communication is an important factor in an FL setting. It is after all how all the training of ML models occur. Despite its' importance there is still limited research done in comparison to other aspects such as privacy, security and resources. For future expectations we hope to see research on this topic to further grow towards making the efforts making them more efficient.

\bibliographystyle{IEEEtran}
\bibliography{mybib}

\hspace{5 cm}


\end{document}